%% file: main.tex
\documentclass[10pt,twocolumn,letterpaper]{article}

\usepackage{wacv}              % To produce the CAMERA-READY version
\usepackage{graphicx}
\usepackage{amsmath}
\usepackage{amssymb}
\usepackage{booktabs}

\usepackage{algorithmicx}
\usepackage{algorithm}
\usepackage{algpseudocode}
\usepackage{lmodern}

\usepackage{hhline}
\usepackage{array}
\usepackage{subcaption}
\usepackage{multirow}
\DeclareMathOperator*{\argmin}{arg\,min}
\DeclareMathOperator*{\argmax}{arg\,max}
\newcolumntype{C}[1]{>{\centering\let\newline\\\arraybackslash\hspace{0pt}}m{#1}}
\let\tb\textbf

\usepackage[pagebackref,breaklinks,colorlinks]{hyperref}

\usepackage[capitalize]{cleveref}
\crefname{section}{Sec.}{Secs.}
\Crefname{section}{Section}{Sections}
\Crefname{table}{Table}{Tables}
\crefname{table}{Tab.}{Tabs.}

\def\wacvPaperID{2407} % *** Enter the WACV Paper ID here
\def\confName{WACV}
\def\confYear{2024}

\begin{document}

\title{Hard-label based Small Query Black-box Adversarial Attack}

\author{Jeonghwan Park, Paul Miller, Niall McLaughlin\\
Queen's University Belfast, United Kingdom\\
{\tt\small \{jpark04, p.miller, n.mclaughlin\}@qub.ac.uk}
%\and
}
\maketitle

\input{sec/0_abstract}
\input{sec/1_introduction}
\input{sec/2_related_work}
\input{sec/3_problem_definition}
\input{sec/4_optimisation_framework}
\input{sec/5_experiments}

\input{sec/6_conclusion}

%%%%%%%%% REFERENCES
{\small
\bibliographystyle{ieee_fullname}
\bibliography{main}
}

\vspace{20pt}

\input{supplemental_material}

\end{document}

%% file: sec/0_abstract.tex
\begin{abstract}
We consider the hard-label based black-box adversarial attack setting which solely observes the target model's predicted class. Most of the attack methods in this setting suffer from impractical number of queries required to achieve a successful attack. One approach to tackle this drawback is utilising the adversarial transferability between white-box surrogate models and black-box target model. However, the majority of the methods adopting this approach are soft-label based to take the full advantage of zeroth-order optimisation. Unlike mainstream methods, we propose a new practical setting of hard-label based attack with an optimisation process guided by a pre-trained surrogate model. Experiments show the proposed method significantly improves the query efficiency of the hard-label based black-box attack across various target model architectures. We find the proposed method achieves approximately 5 times higher attack success rate compared to the benchmarks, especially at the small query budgets as 100 and 250. 
\end{abstract}

%% file: sec/1_introduction.tex
\section{Introduction}
\label{Sec:Introduction}

Deep neural networks (DNNs) have become a most successful back-bone technique adopted in many machine learning applications. Especially in the object classification domain, DNN models can classify objects in images with near human accuracy. However, such models are in general vulnerable to adversarial attacks which use maliciously modified input examples to mislead a target model.~\cite{L-BFGS}. 

Depending on the amount or type of information of target models the attack methods can access, adversarial attack methods are divided into two categories: white-box and black-box attacks. White-box attacks are carried out with full access to necessary information of the target models, such as their weights and structures, to efficiently conduct back-propagation and calculate gradients \cite{FGSM, MI-FGSM}. In the black-box attack settings, however, the attack methods can only monitor the input examples and corresponding output information from the target model \cite{HSJA, BA, NES}. The black-box adversarial attacks are, therefore, considered to be more practical attack methods compared to the white-box attacks. 

Carlini et al.~\cite{ADV_ROBUSTNESS} has emphasised the importancy of black-box adversarial attack to evaluate the robustness or security of applications running with DNN model. The applicable black-box attack method was introduced by Papernot et al.~\cite{SUBSTITUTE} in early 2016. Black-box attacks are carried out without knowledge of the target model, and rely on the target model's output data obtained through model querying. Therefore, such attack methods utilise optimisation processes to approximate gradients to generate adversarial examples. Depending on the method to approximate gradient, Black-box attack methods are broadly categorised into transfer based and query based attacks. Transfer based attacks utilise surrogate models to generate gradients which are directly used as the approximate gradient, since such gradients are likely remain adversarial for the target model due to their transferability~\cite{SUBSTITUTE,TRANSFERABILITY}. Although various methods have been introduced to improve the transferability~\cite{MI-FGSM,VNI-FGSM}, the attack success rate (ASR) is yet to be satisfactory. The reason is that there lacks an adjustment procedure in transfer based attacks when the gradient of the surrogate model points to a non-adversarial region of the target model\cite{P_RGF2}. Query based attacks adopt various gradient-free optimisation processes, which search the optimal point of the functions without using their derivative, to estimate gradients~\cite{ZOO,AUTOZOOM, NES}. Consider a DNN model which classifies an input example into a class. The real-world applications, which deal with the classification problem utilising the DNN, generate a predicted soft-label and/or hard-label as the classification result. According to Galstyan et al.~\cite{SOFT_HARD_LABEL}, a hard label is one assigned to a member of a class where membership is binary, and a soft label is one which has a score (probability or likelihood) attached to it. Based on this definition, the query based attacks subdivided further into soft-label based, and hard-label based attacks. As the target model's classification probability distribution is not observable to the attacks, hard-label based attacks are generally considered to be more challenging task compared with soft-label based attacks. Although query based attack methods generally achieve a higher ASR in comparison with the transfer based attacks, regardless of their categories they typically suffer from a large number of queries required to perform a successful attack. 

In this report a novel method is introduced to tackle the query in-efficiency of the hard-label based black-box attack methods by proposing a black-box attack method called Small-Query Black-Box Attack (SQBA). SQBA integrates the transfer based attack to take advantage of the gradients generated from a surrogate model, and applies gradient-free optimisation introduced in \cite{HSJA}. In summary, the contributions of this report are:
\begin{itemize}
  \item A novel transfer based iterative gradient estimation is proposed to guide gradient direction in the black-box attack settings.
  \item We design a hard-label based black-box attack method, SQBA, which has an optimisation process guided by the proposed transfer based gradient estimation.  
  \item Through the experiments the improved query-efficiency of SQBA attack method is demonstrated in comparison with several state-of-the-art hard-label based black-box attack methods. 
\end{itemize}

%% file: sec/2_related_work.tex
\section{Related works}
\label{Sec:Related_works}

In the black-box adversarial attack settings, the attack methods need to generate adversarial examples without gradient information from the target model, as they have no or limited accessibility to the model. A common choice to tackle this problem is to utilise an approximate gradient instead for generating adversarial examples. 
While the mainstream approach in generating the approximate gradients is to numerically estimate them by the zeroth-order optimisation algorithms\cite{NES,HSJA,SIGN_OPT,BA,BANDIT}, transfer based attacks take an advantage of white-box attack method (surrogate model) to estimate the gradients. Intuition in the transfer based attack methods is that adversarial examples which successfully attack a surrogate model are likely to remain adversarial for the target model due to their transferable characteristics~\cite{TRANSFERABILITY}. 

Tramer et. al. demonstrated in their work~\cite{TRANSFERABLE_ADV} that adversarial examples display a high level of transferability between models, which are trained for the same task, but have different architectures. To exploit this characteristic of the adversarial examples, transfer based attack methods train substitute (surrogate) models to achieve reasonably similar behaviour to the target model, and then apply generic white-box attacks to produce approximate gradients to attack the target model. In this report, however, it is assumed a surrogate model which well clones the target model is given, considering an easy access to the off the shelf pre-trained models, whose network structures are different from the target model.

Adversarial examples generated from the surrogate model are not always transferable to the target model, although many efforts have been made to improve the transferability of the adversarial examples by utilising momentum guidances~\cite{MI-FGSM, VNI-FGSM}, or transforming input examples~\cite{DIM, TIM, SIM}. The failure of transferability is often presented when the classification probabilities of the models are different, and such discrepancy tends to cause disagreement on gradient directions. 

The query based attack methods commonly utilise the iterative-querying approach adopting the zeroth-order optimisation. In details, the approximate gradient is estimated by finite difference~\cite{ZOO,GRAD-ESTIMATE-BA}, random-gradient estimation~\cite{AUTOZOOM} and natural evolution strategy (NES)~\cite{NES, BANDIT}. However, such attack methods suffer from impractically large numbers of queries to achieve successful attack. It is because they typically spend many queries to find the intermediate adversarial examples as the nature of zeroth order optimisation becomes effective if and only if examples lie near the decision boundary of the target model~\cite{BA,SIGN_OPT,HSJA}. 

To overcome this drawback of the query based attack methods, an approach to leverage transfer based and query based methods have been adopted in recent studies. Brunner et al.~\cite{BIASED_BA} integrated a white-box attack method within Boundary Attack (BA) method~\cite{BA}, and used adversarial example originated from the gradient of the surrogate white-box model to guide optimisation process. Yang et al.~\cite{LeBA} improved SimBA~\cite{SIMBA} query based attack and applied on-line surrogate model updating algorithm in which simulates both forward and backward process of target model. Zhou et al.~\cite{ADV-EIGEN} systematically integrated the surrogate model to generate a Jacobian matrix and improved query efficiency adopting optimisation process of SimBA. Dong et al.~\cite{P_RGF2} obtained the gradient information from a surrogate model, and used it to accelerate searching process. Huang et al.~\cite{TREMBA} generated gradients from a surrogate model, and applied NES~\cite{NES} optimisation process to search adversarial examples. 

Most of methods reviewed here utilise the optimisation processes rely on the soft-label output of the target model, whereas \cite{BIASED_BA} focuses on the similar attack settings as SQBA method which uses the hard-label optimisation process. 

%% file: sec/3_problem_definition.tex
\section{Problem Definition}
\label{Sec:Problem_Definision}

Suppose $F(x)$ is a model to deal with a $k$-class object classification problem, and the output of the model is denoted as $f(x) \rightarrow \mathbb{R}^{k}$ to quantify classification probabilities of the input example $x \in \mathbb{R}^{m}$ belonging to the classes. The successful classification is represented as $f_{c^{\dagger}}(x) = \max_{k}[f_{k}(x)]$, where $c^{\dagger}$ is the true class of $x$. 

The goal of the adversarial attack is to find a modified example $\tilde{x}$ such that $f_{\tilde{c}}(\tilde{x}) > f_{c^{\dagger}}(\tilde{x})$ and $\mathcal{D}(x, \tilde{x})$ is small enough, where $\tilde{c}$ is an adversarial class and $\mathcal{D}$ is a dissimilarity metric. In hard-label based black-box attack settings, the attack methods iteratively inquire the target model with queries using purposely modified examples $x'_{t}$ and only observe the predicted classes. The parameters $\theta$ in the target model and the classification probabilities $\{f_{0}(x'_{t}), f_{1}(x'_{t}), ..., f_{k}(x'_{t})\}$ are not accessible to the attack methods.  

%% file: sec/4_optimisation_framework.tex
\section{Optimisation Framework}
\label{Sec:Optimisation_Framework}

The output vector of model $F(x)$ is a probability distribution over the class set $\textbf{k} = \{1, . . . , k\}$. The classification function of $F(x)$ is denoted as $C : \mathbb{R}^{m} \rightarrow \textbf{k}$ which maps input example $x$ to the class with highest probability, and it is defined as~\cite{HSJA}:
\begin{equation}
C(x) := \argmax\limits_{c \in \textbf{k}} [f_{c}(x)]
\label{Eq:Ch3_SQBA_C(x)} 
\end{equation}
\noindent Suppose a function $\mathcal{H}(x)$ is defined for the untargeted attack to divert the true classification $c^{\dagger} = C(x)$ to any unknown class, which is an adversarial class, $\tilde{c} \in \textbf{k}$ subject to $\tilde{c} \neq c^{\dagger}$ as:
\begin{equation}
\mathcal{H}(x + \alpha\mu) := \lim_{\alpha \rightarrow \epsilon} \Bigl( f_{\tilde{c}}(x + \alpha\mu) - \max\limits_{\tilde{c} \neq c}[ f_{c}(x + \alpha\mu) ] \Bigr)
\label{Eq:Ch3_SQBA_H(x)_Definition}
\end{equation}
\noindent where $\mu$ is a perturbation vector generated with an intention to conduct adversarial attack, $\alpha$ is a scaling factor, and $\epsilon \leq 1$ is a small positive value. For the targeted attack, the adversarial class $\tilde{c}$ becomes a designated class. When $\mathcal{H}(x')$ is positive, given modified example $x'$ is clearly adversarial, while negative $\mathcal{H}(x')$ means $x'$ lies in the true class region. Especially when $\mathcal{H}(x')$ is zero, $x'$ is indicated to lie exactly on the decision boundary between two classes, $c^{\dagger}$ and $\tilde{c}$. The examples which fall into this case are called boundary-example~\cite{QEBA}. 

In hard-label based attack setting, however, $\mathcal{H}(x')$ is not a linear function and observed as the Heaviside step function since $f_{c}(x)$ is not accessible. Therefore, $\mathcal{H}(x')$ needs to be re-defined as: 
\begin{equation}
  \mathcal{H}(x')\ := \left.
  \begin{cases}
    \;\;\,1,   \;\; \text{if} \; f_{\tilde{c}}(x') > \max\limits_{\tilde{c} \neq c}[ f_{c}(x') ] \\
   -1,   \;\; \text{otherwise.}      
  \end{cases}
  \right.
\label{Eq:Ch3_SQBA_H(x)_Boolean}
\end{equation}
\noindent where $x' = x + \alpha\mu$. When $\alpha$ is acceptably small, the status change of $\mathcal{H}(x')$ is observed if and only if $x'$ is a direct neighbour of the boundary-example. In the hard-label based attack settings, therefore, it is necessary to keep the distance between $x'_{t}$ and decision boundary close enough in the search phase to guarantee the successful attack with high quality adversarial examples.

As shown in \Cref{Eq:Ch3_SQBA_H(x)_Boolean}, $\mathcal{H}(x')$ is a boolean function, and therefore it is useful to indicate the successful adversarial examples. With the boolean function, the objective of the adversarial attacks can be seen as a process to find an example which satisfies: 
\begin{equation}
\min\limits_{x'} \mathcal{D}(x', x), \quad \text{such that} \; \mathcal{H}(x') = 1
\label{Eq:Ch3_SQBA_Adv_definition}
\end{equation}
\noindent where $\mathcal{D}$ is a distance metric to quantify the amount of applied perturbation vector to modify the given example. Typically $l_{p}$-norm, such that $p \in\{0, 2, \infty\}$, is utilised for this purpose. In this work, $l_{2}$-norm distance metric, $\mathcal{D}(x', x) = \|x' - x\|_{2}$, is chosen. 

Consider the objective of the adversarial attacks defined in \Cref{Eq:Ch3_SQBA_Adv_definition} as an optimisation problem. Two gradients are estimated  in this work to solve such problem, and to effectively control the error in estimations. The first gradient $\nabla \mathcal{H}_{w}(x')$ is estimated via following process:  
\begin{equation}
\nabla \mathcal{H}_{w}(x'_{t}) := \mu_{t}^{i} \quad \text{subject to} \; \min \limits_{\mu_{t}^{i}} \mathcal{D}(x, x'_{t} + \delta_{t} \mu_{t}^{i}) 
\label{Eq:Ch3_SQBA_white_gradient}
\end{equation}
\noindent where $\mu_{t}^{i} = \nabla \mathcal{J}_{S}( x + v_{t}^{i}, \; c^{\dagger} ) $ is a gradient generated from a pre-trained surrogate model $S(x)$, which well clones the classification behaviour of the target model $F(x)$, $v_{t}^{i}$ is a vector to differentiate input example $x$, and $\delta_{t}$ is a scaling factor for the iteration $t$. The gradient $\mu_{t}^{i}$ is calculated through a white-box attack by injecting modified example $(x + v_{t}^{i})$ and true class $c^{\dagger}$ of $x$ into the surrogate model $S(x)$. Specifically, $l_{\infty}$ gradient from \textbf{Dual Gradient Method} (DGM), which is detailed in the Supplemental Material, is used for this purpose. In each iteration $n$ gradients are generated as, $\{\mu_{t}^{0}, \mu_{t}^{1}, ..., \mu_{t}^{n}\}$, and a gradient which can move current intermediate adversarial example $x'_{t}$ to the closest point to the input example $x$ preserving adversariality, is selected. This process is detailed further in section \Cref{Sec:Gradient from surrogate model} of this report. Boolean function $\mathcal{H}(x')$ provides an approximation of the gradient direction, which has been utilised in various black-box attack methods \cite{HSJA, SIGN_OPT, NES, SIMBA}. However, \Cref{Eq:Ch3_SQBA_white_gradient} estimates $\nabla\mathcal{H}_{w}(x'_{t})$ including its likely direction. Therefore, the proposed attack method doesn't apply such approximation to $\nabla\mathcal{H}_{w}(x'_{t})$ for more effective attack process. 
\begin{algorithm}[t]
\footnotesize
{
\caption{SQBA Adversarial Example Computation}
\begin{algorithmic}
%\Input { example $x$, true class $c^{\dagger}$, classifier $f$}
\State $\mbox{\textbf{input:} classifier \textit{f}, example \textit{x}, true class }$ $c^{\dagger},$
\State $\mbox{\qquad\quad\;   surrogate mode \textit{S} }$
\State $\mbox{\textbf{output:} adversarial example }$ $\tilde{x}$
\State $t = 0, \; \beta = 1, \; x'_{t} = \text{sign}\left(\nabla_{S}\mathcal{J}(x, c^{\dagger})\right)$ 
\State $x'_{t+1} = \;$ \Cref{Eq:Ch3_SQBA_binary_search} $\leftarrow x, \; x'_{t}$ 
\While{queries $\leq$ query budget} 
	\State $t = t+1, \; \delta_{t} = 10^{-2}\mathcal{D}(x, x'_{t})$
	\State $\nabla\mathcal{H}_{w} = \;$ \Cref{Eq:Ch3_SQBA_white_gradient} $\leftarrow \delta_{t}$
	\State $\nabla\mathcal{H}_{b} = \;$ \Cref{Eq:Ch3_SQBA_black_gradient} $\leftarrow \delta_{t}$
	\State $g_{t} = \;$ \Cref{Eq:Ch3_SQBA_gradient_update} $\leftarrow \nabla\mathcal{H}_{w}, \nabla\mathcal{H}_{b}, \beta$
	\State $\dot{x}_{t}, \; \beta = \;$ \Cref{Eq:Ch3_SQBA_approch_boundary} $\leftarrow x'_{t}, \; g_{t}$
	\State $x'_{t+1} = \;$ \Cref{Eq:Ch3_SQBA_binary_search} $\leftarrow x, \; \dot{x}_{t}$
\EndWhile \\
$\tilde{x} = x'_{t}$\;
\end{algorithmic}
}
\label{Alg:Ch2_DGM_adv_example_computation}
\end{algorithm}
The gradient estimation algorithm used in HSJA attack method~\cite{HSJA} is adopted in the proposed attack method to estimate the second gradient $\nabla \mathcal{H}_{b}(x'_{t})$. The algorithm generates $p_{t}$ random examples $\mu_{t}^{i}$ which are uniformly distributed, where $\mu_{t}^{i} \in \mathbb{R}^{m}$ and $ i = \{0, 1, ..., p_{t} \}$. At every iteration, the number of random examples is re-calculated. In the proposed method smaller $p_{t} = 10\sqrt{t+1}$, compared with HSJA, is sufficient as the optimisation has already been progressed with $\nabla \mathcal{H}_{w}(x'_{t})$. The second gradient is then estimated from the generated random examples via the Monte Carlo method as:
\begin{equation}
\nabla \mathcal{H}_{b}(x'_{t}) := \frac{1}{p_{t}} \sum^{p_{t}}_{i=1}\mathcal{H}(x'_{t} + \delta_{t} \mu^{i}_{t})\mu^{i}_{t}
\label{Eq:Ch3_SQBA_black_gradient}
\end{equation}
\noindent where $\mu^{i}_{t} \sim \mathcal{N}(O, I)$. $\delta_{t}$ is a scaling factor which is also applied in \Cref{Eq:Ch3_SQBA_white_gradient}, and updated at every iteration as $\delta_{t} = 10^{-2}\mathcal{D}(x, x'_{t})$. With the estimated gradients $\nabla\mathcal{H}_{w}(x'_{t})$ and $\nabla\mathcal{H}_{b}(x'_{t})$, the iterate gradient $g_{t}$ is approximated as:
\begin{equation}
g_{t} \approx \beta_{t} \frac{ \nabla \mathcal{H}_{w}(x'_{t}) } { \| \nabla \mathcal{H}_{w}(x'_{t}) \|_{2} } + (1 - \beta_{t}) \frac{ \nabla \mathcal{H}_{b}(x'_{t}) } { \| \nabla \mathcal{H}_{b}(x'_{t}) \|_{2} }
\label{Eq:Ch3_SQBA_gradient_update}
\end{equation}
\noindent Gradient combining factor $\beta$ is obtained from a boolean function whose output is switched between $\{0, 1\}$. The proposed method begins with $\beta = 1$ meaning $\nabla \mathcal{H}_{w}(x'_{t})$ is solely used to approximate iterate gradients. When $\nabla \mathcal{H}_{w}(x'_{t})$ is indicated to be in a local minima, $\beta$ is then switched to 0 to start using $\nabla \mathcal{H}_{b}(x'_{t})$ instead.  
\begin{figure}[b!]
\centering
\includegraphics[width=2.4in]{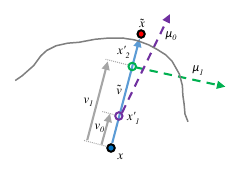}
\caption{Illustration of the process to find iterate example search direction.} 
\label{Fig:Ch3_example_search_direction}
\end{figure}
\begin{figure*}[t]
\centering
\begin{subfigure}{.5\textwidth}
 \centering
 \includegraphics[width=2.2in]{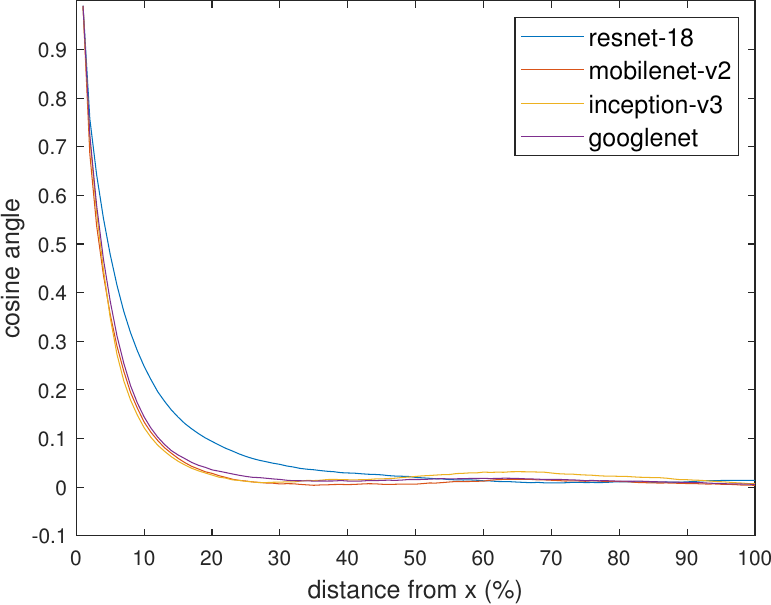}
 \caption{CIFAR-10}
\end{subfigure}%
\begin{subfigure}{.5\textwidth}
 \centering
 \includegraphics[width=2.2in]{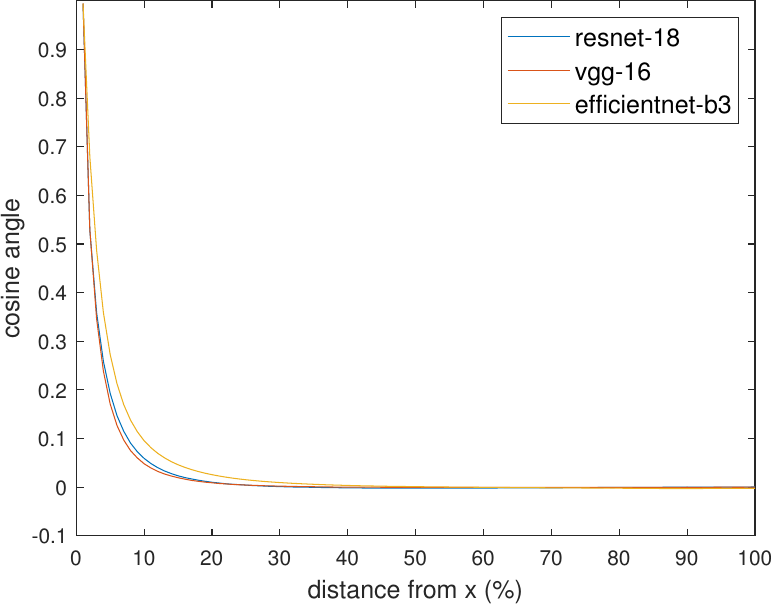}
 \caption{ImageNet}
\end{subfigure}
\caption{Change of Angle of the gradient vectors from surrogate model.}
\label{Fig:Ch3_SQBA_gradient_angle}
\end{figure*}
Consider an iterative optimisation process which is given access to the approximate gradient $g_{t}$ with an intermediate adversarial example $x'_{t}$. The process performs an update as $\dot{x}_{t} = x'_{t} + \alpha g_{t}$. The optimisation problem in \Cref{Eq:Ch3_SQBA_Adv_definition} becomes meaningful when $\dot{x}_{t}$ is a close neighbour to the boundary-example. Therefore, a tuning process is necessary to move $\dot{x}_{t}$ towards the boundary such as: 
\begin{equation}
\dot{x}_{t} = \min\limits_{\alpha} \left( x'_{t} + \alpha g_{t} \right), \quad \text{such that} \; \mathcal{H}(\dot{x}_{t}) = 1 
\label{Eq:Ch3_SQBA_approch_boundary}
\end{equation}
\noindent where $\alpha$ is a line search parameter. The tuning process shall guarantee obtained $\dot{x}_{t}$ to satisfy $\mathcal{H}(\dot{x}_{t}) = 1$ and also $\min \mathcal{D}(x'_{t}, \dot{x}_{t})$, so that the next iterate example lies on the boundary. It can be noticed when an approximate gradient $g_{t}$ gets into a local or global minima, the line search parameter $\alpha$ likely becomes small to find $\dot{x}_{t}$. Therefore, $\alpha$ is used to switch the boolean function output $\beta$ in \Cref{Eq:Ch3_SQBA_gradient_update}. In practice $\alpha \leq 1.0$ is used for this purpose. 

The final step of the iterative optimisation is increasing the correlation between input example $x$ and calculated intermediate adversarial example $\dot{x}_{t}$. A binary search algorithm is applied for the improved correlation as 
\begin{equation}
x'_{t+1} = \lim\limits_{\zeta \rightarrow 0} \left( \zeta x + (1 - \zeta) \dot{x}_{t} \right), \; \text{such that} \; \mathcal{H}(x'_{t+1}) = 1 
\label{Eq:Ch3_SQBA_binary_search}
\end{equation}
\noindent The process shown in \Cref{Eq:Ch3_SQBA_binary_search} is a simple and greedy search algorithm, however it is useful enough to be adopted in many hard-label based methods \cite{HSJA,SIGN_OPT}. 

%-------------------------------------------------------------------------
\subsection{Gradient from surrogate model} 
\label{Sec:Gradient from surrogate model}

The transferability of gradients generated from the surrogate model is mitigated when the decision boundary of the surrogate model does not closely match the target model. However, even when this is the case, the decision boundaries of both models may still be reasonably nearby \cite{BIASED_BA}. To address this drawback, the proposed SQBA attack utilises the multi-gradient method. Let $\tilde{x}$ be an adversarial example generated from surrogate model $S(x)$, and $\tilde{v} = \nabla_{S} \mathcal{J}(x, c^{\dagger})$ be a perturbation vector to transpose the input example $x$ to $\tilde{x}$. The examples that lie on $\tilde{v}$ are calculated by $x'_{i} = x + \eta_{i} \tilde{v}$, where $\eta_{i} \leq 1$ is a positive scaling factor which configures the distance from input example, such as $x'_{i} = x$ subject to $\eta = 0$, and $x'_{i} = \tilde{x}$ subject to $\eta_{i} = 1$. As illustrated in \Cref{Fig:Ch3_example_search_direction}, a set of gradients then can be generated from $S(x'_{i})$ as: 
\begin{equation}
\mu_{i} = \nabla_{S} \mathcal{J}(x'_{i}, c^{\dagger}), \quad \text{where} \; i = 0, 1, ..., n
\label{Eq:Ch3_SQBA_gradient_from_surrogate}
\end{equation}
\begin{table}[!b]
%\begin{minipage}{\columnwidth}
%\begin{center}
\centering
\footnotesize
{
\begin{tabular}{C{1.6cm}|C{2.5cm}|C{2.5cm}}
	\toprule			
	%\multicolumn{3}{c} {Model's Accuracy} \\
	%\midrule		
	{Dataset} & {Model}  & {Accuracy} \\
	%\hline	
	\midrule
	\multirow{4}{*}{CIFAR-10} & {ResNet-18}                      & 93.07\%  \\			
	%\cline{2-3}
	                          & {MobileNet-V2}                   & 93.90\%      \\
	%\cline{2-3}
							  & {Inception-V3}                   & 93.74\%    \\
	%\cline{2-3}
	                          & {GoogLeNet}                      & 92.96\%       \\			
	\hline	
	\multirow{3}{*}{ImageNet} & {EfficientNet-b3}                & 83.58\%       \\			
	%\cline{2-3}
	                          & {VGG-16}                         & 78.98\%          \\
	%\cline{2-3}
							  & ResNet-18                        & 76.74\% \\		  
	\bottomrule                        
\end{tabular}
\caption{List of DNN Models and Classification Accuracies } 
\label{Tab:Ch3_SQBA_DNN_accuracy}  
%\end{center}
%\end{minipage}

\bigskip
	
%\begin{minipage}{\columnwidth}
%\begin{center}
\begin{tabular}{C{1.6cm}|C{2.5cm}|C{2.5cm}}
	\toprule			
	%\multicolumn{3}{c} {Target and Surrogate models} \\
	%\midrule		
	{Dataset} & {Target Model} & {Surrogate Model} \\
	%\hline
	\midrule	
	\multirow{4}{*}{CIFAR-10} & \multirow{2}{*}{ResNet-18}       & {MobileNet-V2}    \\			
	%\cline{3-3}
	                          &                                  & {GoogLeNet}       \\
	\cline{2-3}
							  & \multirow{2}{*}{MobileNet-V2}    & {Inception-V3}    \\
	%\cline{3-3}
	                          &                                  & {ResNet-18}       \\		
	\hline	
	\multirow{4}{*}{ImageNet} & \multirow{2}{*}{EfficientNet-b3} & {ResNet-18}       \\			
	%\cline{3-3}
	                          &                                  & {VGG-16}          \\
	\cline{2-3}
							  & \multirow{2}{*}{VGG-16}          & {EfficientNet-b3} \\
	%\cline{3-3}
	                          &                                  & {ResNet-18}       \\           
	\bottomrule                        
\end{tabular}
%\end{center}
%\end{minipage}
}
\caption{Target and Surrogate models mapping}	
\label{Tab:Ch3_SQBA_DNN_list}
\end{table}

\noindent Many works have proven that the optimal search direction in hard-label based attack is perpendicular to  $\tilde{v}$ \cite{BA, HSJA}. Suppose a function $\mathcal{A}(\bar{a}_{1}, \bar{a}_{2})$ is defined to derive the angle between two vectors as $\mathcal{A}(\bar{a}_{1}, \bar{a}_{2}) := \cos \angle (\bar{a}_{1}, \bar{a}_{2})$, where $\cos \angle (\bar{a}_{1}, \bar{a}_{2}) = (\bar{a}_{1} \cdot \bar{a}_{2}) / \|\bar{a}_{1}\|_{2} \|\bar{a}_{2}\|_{2}$.  The angle between $\tilde{v}$ and a gradient vector $\mu_{i}$ is then calculated with $\mathcal{A}(\tilde{v}, \mu_{i})$. 

To exam the direction of gradients of the set of examples $x'_{i}, \; i = 0, 1, ..., n$, dedicated 500 examples to each CIFAR-10 and ImageNet-100 datasets were taken, and average direction of the gradients from various DNN models were plotted in \Cref{Fig:Ch3_SQBA_gradient_angle}. As one can expect, when $\eta_{i}$ is small, in other words, $x'_{i}$ is close to input example $x$, the direction of gradient vector $\mu_{i}$ is near parallel to $\tilde{v}$ as $\mathcal{A}(\tilde{v}, \mu_{i}) \approx 1$. The direction of $\mu_{i}$, however, tends to be exponentially changed even at a small increment in $\eta$, and rapidly saturated to the perpendicular direction to $\tilde{v}$ as $\mathcal{A}(\tilde{v}, \mu_{i}) \approx 0$. Regardless of DNN models or dataset, the similar pattern of changes is observed that is an example which lies at $\eta_{i} \approx 0.2$, and afterwards their gradients are almost saturated to $\mathcal{A}(\tilde{v}, \mu_{i}) \approx 0$. The iterate approximate gradient, therefore, can be chosen from the calculated gradient vectors such that $\eta_{i} \geq 0.2$ and $\mathcal{H}(\tilde{x} + 2\delta \mu_{i}) = 1$ as:
\begin{equation}
\nabla\mathcal{H}_{w}(\tilde{x}) = \mu_{i} \quad \text{where} \; i = \argmin \limits_{\mathcal{D}} [\mathcal{D}(x, \tilde{x} + 2\delta_{t}\mu_{i} )] 
\label{Eq:Ch3_SQBA_select_gradient_from_surrogate}
\end{equation}
\begin{table*}[!t]
\centering
\footnotesize
{
\begin{tabular}{p{1.5cm}|p{4.0cm}|C{1.5cm}|C{1.5cm}|C{1.5cm}|C{1.5cm}|C{1.5cm}}
	\toprule			
	\multicolumn{7}{c} {CIFAR-10: ResNet-18 at $\rho(\tilde{x}) \leq 0.10$} \\
	\midrule
	\multicolumn{2}{c|} {Query Budget}    
				                              & 100         &  250        &  500        &  750        & 1000   \\
	\hline
	%\multirow{5}{*}{Attack} 
	\multirow{5}{*} {\parbox{5cm}{ Attack \\ Methods}}
    & {HSJA~\cite{HSJA}}                      &  1.2\%      &  9.7\%      & 33.2\%      & 46.9\%      & 60.5\%   \\
    & {BA~\cite{BA}}                          &  2.9\%      &  3.7\%      &  4.0\%      &  4.2\%      &  4.5\%   \\
	& {Sign-Opt~\cite{SIGN_OPT}}              &  8.5\%      & 10.5\%      & 11.4\%      & 24.5\%      & 32.1\%   \\                                
	& {Biased-BA(GoogLeNet)~\cite{BIASED_BA}} &  3.0\%      &  5.5\%      & 13.7\%      & 19.2\%      & 30.3\%   \\
	& {SQBA(MobileNet-V2)}                    & \tb{38.4\%} & \tb{52.1\%} & \tb{66.1\%} & \tb{74.7\%} & \tb{82.0\%} \\	
	& {SQBA(GoogLeNet)}                       & 28.1\%      & 40.4\%      & 60.5\%      & 73.0\%      & 79.2\%   \\		                        		                        			                        
	\bottomrule                        
\end{tabular}
\caption{Query Budgets and ASR of hard-label based black-box attacks on ResNet-18 trained on CIFAR-10} 
\label{Tab:Ch3_SQBA_experiment_cifar10_resnet_18}
%\end{center}
%\end{minipage}
%
\bigskip
%
%\begin{minipage}
%\begin{center}
\centering
\begin{tabular}{p{1.5cm}|p{4.0cm}|C{1.5cm}|C{1.5cm}|C{1.5cm}|C{1.5cm}|C{1.5cm}}
	\toprule			
	\multicolumn{7}{c} {CIFAR-10: MobileNet-V2 at $\rho(\tilde{x}) \leq 0.10$} \\
	\toprule
	\multicolumn{2}{c|} {Query Budget}    
		       		                                         & 100         & 250         & 500         & 750         & 1000   \\
	\hline
	%\multirow{5}{*}{Attack} 
	\multirow{5}{*} {\parbox{5cm}{ Attack \\ Methods}}
			    & {HSJA~\cite{HSJA}}                         &  3.9\%      & 23.4\%      & 67.6\%      & 78.4\%      & 85.6\%   \\
	%\cline{2-7}
			    & {BA~\cite{BA}}                             &  9.3\%      & 10.0\%      & 11.9\%      & 13.1\%      & 16.1\%   \\
	%\cline{2-7}
			    & {Sign-Opt~\cite{SIGN_OPT}}                 & 26.8\%      & 32.0\%      & 37.3\%      & 61.3\%      & 68.9\%   \\                       
	%\cline{2-7}              			                        
			    & {Biased-BA(Inception-V3)~\cite{BIASED_BA}} & 11.9\%      & 28.0\%      & 56.3\%      & 68.0\%      & 73.8\%   \\
	%\cline{2-7}              
			    & {SQBA(Inception-V3)}                       & 71.9\%      & 80.2\%      & 89.4\%      & 91.9\%      & 95.6\%      \\	
	%\cline{2-7}              
			    & {SQBA(ResNet-18)}                          & \tb{72.0\%} & \tb{83.6\%} & \tb{91.1\%} & \tb{95.6\%} & \tb{97.6\%} \\		                        		                        			                        
	\bottomrule                        
\end{tabular}
%\end{center}
%\end{minipage}
}
\caption{Query Budgets and ASR of hard-label based black-box attacks on MobileNet-V2 trained on CIFAR-10} 
\label{Tab:Ch3_SQBA_experiment_cifar10_mobilenet_v2}

\end{table*}

%% file: sec/5_experiments.tex
\section{Experiments}
\label{Sec:Experiments}

Experimental analysis carried out to evaluate SQBA attack method is detailed here. The efficiency of SQBA attack was compared with several state-of-the-art attack methods which utilise the hard-label based adversarial attack setting. 

\subsection{Benchmarks}
\label{Sec:Benchmarks}
SQBA attack method was compared with four state-of-the-art black-box attack methods: HSJA Attack~\cite{HSJA}, Sign-Opt Attack~\cite{SIGN_OPT}, BA Attack~\cite{BA} and Biased-BA Attack~\cite{BIASED_BA}. These attack methods commonly fall into the hard-label based black-box adversarial attack setting as they solely observe the predicted class against the given example. Especially Biased-BA is taking the similar approach as SQBA which embeds a transfer based algorithm within the base line hard-label based attack method. Model parameters or model output logits are not accessible to the attack methods used in the experiments. All the benchmark methods were configured to perform the untargeted attack scenario. In the untargeted attack scenario, an attack becomes successful if the target model predicts a different class from the true class of the given example. Note that Biased-BA attack was initialised with a random noise instead of a target example to conduct such an attack scenario. Attack strategies of the benchmark methods commonly have three stages: First, they find an initial adversarial example (starting point); Second, they iteratively search a path towards a further optimised example, which is closer to the original example but still maintains the adversariality; and finally, they stop searching when the optimisation process satisfies stopping criteria which is given the query budget. 

\subsection{Datasets and Models}
\label{Sec:Data_and_Models}
\begin{table*}[t!]
\centering
\footnotesize
{
\begin{tabular}{p{1.5cm}|p{3.5cm}|C{1.5cm}|C{1.5cm}|C{1.5cm}|C{1.5cm}|C{1.5cm}}
	\toprule			
	\multicolumn{7}{c} {ImageNet: EfficientNet-b3 at $\rho(\tilde{x}) \leq 0.10$} \\
	\midrule
	\multicolumn{2}{c|} {Query Budget}    
				                           & 100         &  250        &  500        &  750        & 1000   \\
	\hline
	\multirow{5}{*}{\parbox{5cm}{ Attack \\ Methods}} 
    & {HSJA~\cite{HSJA}}                   &  1.7\%      &  5.2\%      & 12.0\%      & 19.3\%      & 24.6\%   \\
	& {BA~\cite{BA}}                       &  2.7\%      &  3.0\%      &  3.0\%      &  3.2\%      &  3.4\%   \\
	& {Sign-Opt~\cite{SIGN_OPT}}           &  5.5\%      &  7.5\%      &  7.2\%      &  7.5\%      &  7.3\%   \\                       
	& {Biased-BA(VGG-16)~\cite{BIASED_BA}} &  2.0\%      &  3.4\%      &  7.5\%      & 15.6\%      & 21.6\%   \\
	& {SQBA(ResNet-18)}                    & \tb{33.8\%} & \tb{38.2\%} & \tb{46.8\%} & \tb{49.4\%} & \tb{55.2\%} \\	
	& {SQBA(VGG-16)}                       & 30.4\%      & 37.4\%      & 45.0\%      & 48.7\%      & 53.2\%   \\	
	\bottomrule                        
\end{tabular}
%\end{center}
\caption{Query Budgets and ASR of hard-label based black-box attacks on EfficientNet-b3 trained on ImageNet} 
\label{Tab:Ch3_SQBA_experiment_imagenet_efficientnet}
\bigskip\centering
\begin{tabular}{p{1.5cm}|p{3.5cm}|C{1.5cm}|C{1.5cm}|C{1.5cm}|C{1.5cm}|C{1.5cm}}
	\toprule			
	\multicolumn{7}{c} {ImageNet: VGG-16 at $\rho(\tilde{x}) \leq 0.10$} \\
	\midrule
	\multicolumn{2}{c|} {Query Budget}    
		       		                         & 100         & 250         & 500         & 750         & 1000   \\
	\hline
	\multirow{5}{*} {\parbox{5cm}{ Attack \\ Methods}}
	& {HSJA~\cite{HSJA}}                     &  2.2\%      & 10.2\%      & 24.0\%      & 31.2\%      & 40.8\%   \\
    & {BA~\cite{BA}}                         &  5.0\%      &  6.0\%      &  6.8\%      &  6.2\%      &  9.0\%   \\
	& {Sign-Opt~\cite{SIGN_OPT}}             & 11.4\%      & 13.6\%      & 14.4\%      & 15.8\%      & 15.7\%   \\                       
	& {Biased-BA(ResNet-18)~\cite{BIASED_BA}}&  6.3\%      &  9.0\%      & 22.7\%      & 34.4\%      & 40.6\%   \\
	& {SQBA(Inception-V3)}                   & 57.6\%      & 62.8\%      & 68.4\%      & 72.4\%      & 75.8\%      \\	
	& {SQBA(ResNet-18)}                      & \tb{65.5\%} & \tb{69.2\%} & \tb{73.3\%} & \tb{78.0\%} & \tb{80.9\%} \\		                       
	\bottomrule                        
\end{tabular}
}
\caption{Query Budgets and ASR of hard-label based black-box attacks on VGG-16 trained on ImageNet}
\label{Tab:Ch3_SQBA_experiment_imagenet_vgg_16}
\end{table*} 
The experiments were conducted over two standard image datasets: CIFAR-10~\cite{CIFAR-10} and ImageNet-100~\cite{IMAGENET-1000}. CIFAR-10 dataset is composed of RGB image examples in the $3 \times 32 \times 32$ dimensional spaces. It has 10 classes with 6,000 examples per class. ImageNet-100 is a subset of ImageNet-1k dataset. It consists of random 100 classes out of 1K classes of the full dataset, and each class has 1,350 examples. All the models used in the experiments are listed in \Cref{Tab:Ch3_SQBA_DNN_accuracy}. The models were off the shelf, which were not re-trained nor modified, obtained from publicly available libraries, such as Zenodo~\cite{CIFAR10_MODELS} for CIFAR-10 and Pytorch torchvision library~\cite{IMAGENET_MODELS} for ImageNet-1K models. All the examples were transformed as instructed by the libraries. For example, examples in ImageNet-100 dataset were rescaled to $3 \times 256 \times 256$ dimensional space, and centre cropped to $3 \times 224 \times 224$ dimensional RGB examples. From the validation split of each dataset, dedicated subsets were created with the examples correctly classified by the target models to avoid artificial inflation of the success rate. 1000 examples were randomly down sampled further from the subsets to be used in the evaluation. As the objective dataset, ImageNet-100, is a direct subset of ImageNet-1K dataset, DNN models trained on ImageNet-1K dataset are naturally ready for the objective dataset. In the rest of the report, ImageNet-100 dataset is presented as ImageNet unless otherwise specified. \\ 
\indent For fair and easy comparison, the DNN models, which are publicly well known and pre-trained with the datasets described above, were chosen for the experiments. Specifically, for the experiments with CIFAR-10 dataset, two DNN models, ResNet-18 and MobileNet-V2 were used as the target models. SQBA attack method conducted two attacks on each target model utilising different surrogate models, whose architectures are different from others, to demonstrate the generalisation of the method. To attack ResNet-18 target model, MobileNet-V2 and GoogLeNet were chosen for the surrogate model, while Inception-V3 and ResNet-18 were used to attack MobileNet-V2. For the experiments with ImageNet dataset, VGG-16 and EfficientNet-b3 were attacked by the methods. SQBA attack method utilised ResNet-18 and EfficientNet-b3 models to perform attacks on VGG-16 target model. ResNet-18 and VGG-16 models were used as the surrogate models to attack EfficientNet-b3 target model. Each group of models used in the experiments are detailed in Table~\ref{Tab:Ch3_SQBA_DNN_list}.  

\subsection{Adversarial Attack Budgets}
\label{Sec:Adversarial_Attack_Budgets}
In order to generate meaningful adversarial examples, it is necessary to impose reasonable constraints to the attack. An unconstrained attack perhaps achieve a successful attack by, for example, disturbing target model with tremendous number of queries, or introducing large and brutal perturbations to an input example that would alter its semantics~\cite{ADV_ROBUSTNESS}. Since such attacks are not considered to satisfy the goal of adversary, which is an adversarial example shall have human-imperceptible perturbation, two restrictions were introduced in the experiments. In the rest of this report, the restrictions are called attack budgets.

The first attack budget used in order to quantify the performance of the attack methods is a metric used in~\cite{DEEPFOOL} as:
\begin{equation}
\rho(\tilde{x}) := \frac{ \| \mu \|_{2} }{ \| x \|_{2} }
\label{Eq:Ch3_SQBA_Metric_1}
\end{equation}
\noindent where $\mu$ is the perturbation vector added to the original example $x$ to achieve a successful adversarial attack. This metric was originally used to measure the average robustness. In this report, however, $\rho(\tilde{x})$ is used as the perturbation budget to determine the successfulness of an individual attack. $\rho(\tilde{x})$ in fact indicates the distance between $x$ and $\tilde{x}$, therefore the lower value means the higher robustness. 

\begin{table*}[t!]
\footnotesize
{
\centering
\begin{tabular}{p{1.0cm}|p{3.5cm}|C{1.1cm}|C{1.1cm}}
	\toprule			
	\multicolumn{4}{c} {CIFAR-10: ResNet-18-AT at $\rho(\tilde{x}) \leq 0.10$} \\
	\midrule
	\multicolumn{2}{c|} {Query Budget}    
                    	               		               & 750         & 1000     \\
	\hline
	\multirow{4}{*}{\parbox{5cm}{ Attack \\ Methods}}  
			    & {HSJA~\cite{HSJA}}                                   & 16.5\%      & 22.1\%   \\		   
	%\cline{2-7}
			    & {Sign-Opt~\cite{SIGN_OPT}}                               &  8.6\%      &  9.5\%   \\                       
	%\cline{2-7}              			                        
			    & {Biased-BA(GoogLeNet)~\cite{BIASED_BA}} &  3.0\%      &  5.9\%   \\
	%\cline{2-7}              
			    & {SQBA(GoogLeNet)}     & \tb{18.0\%} & \tb{22.5\%} \\
	\bottomrule                        
\end{tabular}
\quad
\begin{tabular}{p{1.0cm}|p{3.8cm}|C{1.1cm}|C{1.1cm}}
	\toprule			
	\multicolumn{4}{c} {CIFAR-10: MobileNet-V2-AT at $\rho(\tilde{x}) \leq 0.10$} \\
	\midrule
	\multicolumn{2}{c|} {Query Budget}    
				                                              & 750         & 1000     \\
	\hline
	\multirow{4}{*}{\parbox{5cm}{ Attack \\ Methods}}  
			    & {HSJA~\cite{HSJA}}                                      & 24.5\%      & 27.6\%   \\		   
	%\cline{2-7}
			    & {Sign-Opt~\cite{SIGN_OPT}}                                  & 12.1\%      & 13.5\%   \\                       
	%\cline{2-7}              			                        
			    & {Biased-BA(Inception-V3)~\cite{BIASED_BA}}& 10.5\%      & 25.1\%   \\
	%\cline{2-7}              
			    & {SQBA(Inception-V3)}     & \tb{25.2\%} & \tb{31.9\%} \\
	\bottomrule                        
\end{tabular}
}
\caption{Comparison of black-box attacks on defended models for CIFAR-10 dataset} 
\label{Tab:Ch3_SQBA_experiment_robust_model}
\end{table*} 

As the second attack budget, the number of queries that an attack method can use is also restricted by the five different budgets such as $\{$100, 250, 500, 750, 1000$\}$. The highest allowed number of queries 1000 may seem to be in-efficient. However, some object classification applications which utilise modern techniques such as GPU or FPGA accelerated systems are already capable of processing more than 1000 images per second~\cite{GPU_BENCHMARK}. 

%----------------------------------------------------------------------------------------- 
\subsection{Attacks on CIFAR-10 Dataset}
\label{Sec:Attacks_on_CIFAR10_Dataset}

In this section the results of black-box adversarial attacks with CIFAR-10 dataset are reported. Pre-trained ResNet-18 and MobileNet-V2, which achieves 93.07\% and 93.90\% classification accuracies respectively, were chosen for the target models. All the attacks were limited with the perturbation budget $\rho(\tilde{x}) < 0.1$, and to at most the five query budgets as described in \Cref{Sec:Adversarial_Attack_Budgets}. Any successful attack achieved with queries or perturbation which exceeded the budgets was reported as a failed attack. 

As shown in \Cref{Tab:Ch3_SQBA_DNN_list}, Two SQBA with different surrogate models performed black-box attacks to each target model, and their performance was compared with four state-of-the-art hard-label based attack methods listed in \Cref{Sec:Benchmarks}. 
The attack results over ResNet-18 target model is presented in \Cref{Tab:Ch3_SQBA_experiment_cifar10_resnet_18}. SQBA with both MobileNet-V2 and GoogLeNet surrogate models outperformed benchmark methods across all the query budgets, specifically SQBA with MobileNet-V2 showed the best attack performance achieving 82.0\% ASR at 1000 query budget. Especially in the very small query scenarios, where queries are limited with 100 and 250 at most, SQBA attacks performed approximately 4 and 5 times higher ASR respectively compared with the best performing benchmark. \Cref{Tab:Ch3_SQBA_experiment_cifar10_mobilenet_v2} details attack performances on MobileNet-V2 target model. SQBA attacks also outperformed benchmarks in this experiment. At 100 query budget, both SQBA attacks with Inception-V3 and ResNet-18 surrogate models achieved over 70.0\% ASR while other bench mark methods didn't show meaningful attack performances yet. 

\subsection{Attacks on ImageNet Dataset}
This section details the performances of black-box adversarial attacks conducted against target-models trained on ImageNet dataset. For the experiments, pre-trained EfficientNet-b3 and VGG-16 DNN models were chosen for the target models, and they respectively achieve 83.58\% and 78.98\% classification accuracies over ImageNet-100 dataset. Surrogate models used in SQBA attacks are detailed in \Cref{Tab:Ch3_SQBA_DNN_list}. The benchmark methods and attack budgets used in CIFAR-10 experiments were also applied in this series of experiments to evaluate the performance of the SQBA methods. 

\Cref{Tab:Ch3_SQBA_experiment_imagenet_efficientnet} and \Cref{Tab:Ch3_SQBA_experiment_imagenet_vgg_16} show the performances of the attack methods over ImageNet models, EfficientNet-b3 and VGG-16 respectively. While it is a general observation that SQBA attacks outperformed all other methods, in detail SQBA attacks achieved approximately 5 times higher ASR compared with the benchmarks at small query budgets 100 and 250, interestingly all methods struggled to achieve successful attacks to EfficientNet-b3 model. At 1000 query budget, HSJA~\cite{HSJA} performed 40\% degraded ASR compared to the same attack to VGG-16 target model, while SQBA with ResNet-18 achieved 32\% degraded ASR compared with the attack on VGG-16. 

\subsection{Attacks on Defended models}
Attack methods were further evaluated with the defended models. ResNet-18-AT and MobileNet-V2-AT models have identical architectures to the original models used in \Cref{Sec:Attacks_on_CIFAR10_Dataset}. They were trained with CIFAR-10 dataset using the PGD adversarial training method~\cite{PGD_AT} to achieve 67.24\% and 68.49\% standard classification accuracies respectively. \Cref{Tab:Ch3_SQBA_experiment_robust_model} show the comparison of black-box attacks on both defended models. In these experiments, two query budgets, 750 and 1000, were considered, and BA attack method which didn't show meaningful ASR in the previous experiments was not included. GoogLeNet and Inception-V3 were chosen for the surrogate models to attack ResNet-18-AT and MobileNet-V2-AT respectively. It can be observed that the defended models successfully reduced the ASR of all the attack methods to approximately 70.0\% less than their attacks on the original models. Although SQBA attacks showed the improved performances compared with the benchmarks, their performance was also heavily degraded achieving 22.5\% on ResNet-18-AT and 31.9\% on MobileNet-V2-AT target models at 1000 query budget, which were 71.5\% and 66.6\% respectively less ASR compared with the same attacks on the original models.

%% file: sec/6_conclusion.tex
\section{Conclusion}
SQBA method which combines transfer based and hard-label based adversarial attacks was introduced. SQBA is designed to generate likely adversarial examples for an unknown classifier whose predicted class is only accessible. The method takes an advantage of the transfer based algorithm for the rapid search of an initial convergence point which is followed by the guided optimisation procedure. Through the experiments it was demonstrated that SQBA is capable of achieving higher ASR with smaller query budgets compared with the benchmarks. One drawback of black-box attack methods which utilise transfer based algorithms degrades the performance when gradients from the transfer based algorithms disagree with the target model, and SQBA is not immune from this problem. Future work may focus on improving the adversarial transferability in the SQBA method to address this issue. 

%% file: supplemental_material.tex
\section{Supplemental Material}

\subsection{Dual Gradient Attack Method (DGM)}
This supplemental material describes Dual Gradient white-box attack Method(DGM) which is adopted in SQBA attack method. DGM uses dual gradient vectors to effectively find perturbations. The procedure of DGM attack consists of two stages: (1) Generating adversarial perturbation, and (2) Fine-tuning generated perturbation. 
 
\subsubsection{Adversarial Perturbation}
An input example $x \in \mathbb{R}^{m}$ to be estimated by the $k$-class classifier $F(x)$ can be seen as a data point located in a convex polyhedron \cite{DEEPFOOL}, whose faces represent classes $c_{i}$, where $i = 0, 1, ..., k$. The orthogonal distance from a class $c_{i}$ to the data point $x$ denotes the quantified classification probability, and it is expressed as $\Delta(x; c_{i})$. The decision boundary of the classifier can be seen as an affine linear equation also in the polyhedron. Since the primary objective of adversarial attack is to modify input example $x$ to mislead a target model, the true class $c^{\dagger}$ and an adversarial class $\tilde{c}$ need to be considered. \\

\noindent The orthogonal distance between $c^{\dagger}$ and $x$ is smaller than other classes in the polyhedron as $\Delta(x; c^{\dagger}) < \Delta(x; c^{i}), \; \forall \; c^{\dagger} \neq c^{i} $. To mislead the target model, therefore, $\Delta(x; c^{\dagger})$ needs to be increased by moving example $x$ sufficiently to an adversarial region where belongs to another class $c^{i}$. DGM searchs the optimal path to transport $x$ by utilising two vectors, $g^{-}$ and $g^{+}$, which are calculated with respect to the true class $c^{\dagger}$ and a potential adversarial class $c^{i}$ respectively. The vector $g^{-}$ has the negative direction to the class, therefore, it moves $x$ away from the true class. In contrast the positive directional vector $g^{+}$ transports $x$ closer to the adversarial class. The gradient vectors are calculated as: 
\begin{equation}
  g^{+/-} = \left.
  \begin{cases}
    l_{2} \;\;   : \nabla_{F} f_{c}(x)/\mbox{max}\left( \nabla_{F} f_{c}(x) \right)\\
    l_{\infty}\, : \mbox{sign}\left( \nabla_{F} f_{c}(x) \right)     
  \end{cases}
  \right.
\label{Eq:Ch2_DGM_gradient}
\end{equation}
\noindent where $\nabla_{F} f_{c}(x)$ is a gradient vector obtained from the backward process of $F(x)$ with input example $x$ and associated class $c$. DGM iteratively conducts such process to efficiently find the optimal direction of adversarial perturbation vector $\mu_{t}$ as:
\begin{equation}
\mu_{t} = (-\alpha_{t}) g^{-}_{t} + (1-\alpha_{t})g^{+}_{t}
\label{Eq:Ch2_DGM_perturbation_vector}
\end{equation}
\noindent where $\alpha_{t} \in [0, 0.3]$ is a penalty parameter applied to both directional vectors. The penalty parameter is used to help the stable convergence in searching an optimal vector $\mu_{t}$, and updated in every iteration as: 
\begin{equation}
\alpha_{t} = \min\left( 1 / e^{ 4\lambda },\; 0.3 \right), \quad \text{where} \; \lambda = \frac{ f_{c^{i}_{t}}(x'_{t}) }{ \left( f_{c^{\dagger}}(x'_{t}) + f_{c^{i}_{t}}(x'_{t}) \right) }
\label{Eq:Ch2_DGM_penalty_parameter}
\end{equation}
\noindent Finally an intermediate adversarial example is calculated with a scaling factor $\epsilon \leq 1$, which is a small positive value as 
\begin{equation}
x'_{t+1} = \text{clamp} \left( x'_{t} + \epsilon \mu_{t} \right)\big|_{[\text{min}(x), \text{max}(x)]} .
\label{Eq:Ch2_DGM_intermediate_adv}
\end{equation}
\noindent Algorithm~\ref{Alg:Ch2_DGM_adv_example_computation} outlines the process to find an adversarial example with DGM method.
 
% Algorithm
\begin{algorithm}[t]
\begin{algorithmic}
\State $\mbox{\textbf{input:} example $x$, true class $ c^{\dagger}$, classifier } f$
\State $\mbox{\textbf{output:} adversarial example $\tilde{x}$}$
%$t = 0$, $x'_{t} = x$ \;
\While {TRUE} 
	\State $\{c^{0}_{t}, c^{1}_{t}, ..\} = \;$ Sort$(f(x'_{t}),\; descend)$
	\State $\mbox{\textbf{if} $ c^{0}_{t} \neq c^{\dagger}$ \textbf{then} Break; \textbf{end}}$
	\State $\tilde{c} = c^{1}_{t}$
	\State $\alpha_{t} = \;$ Equation \eqref{Eq:Ch2_DGM_penalty_parameter} $\leftarrow \tilde{c}, c^{\dagger}$
	\State $g^{+}_{t} = \;$ Equation \eqref{Eq:Ch2_DGM_gradient} $\leftarrow \tilde{c}$
	\State $g^{-}_{t} = \;$ Equation \eqref{Eq:Ch2_DGM_gradient} $\leftarrow c^{\dagger}$
	\State $\mu_{t} = \;$ Equation \eqref{Eq:Ch2_DGM_perturbation_vector}
	\State $x'_{t+1} = \;$ Equation \eqref{Eq:Ch2_DGM_intermediate_adv}
	\State $t = t+1$\;
\EndWhile \\
$\tilde{x} = \;$ Tune$(x'_{t})$\;
\end{algorithmic}
\caption{DGM Adversarial Example Computation}
\label{Alg:Ch2_DGM_adv_example_computation}
\end{algorithm}
%
% Algorithm
\begin{algorithm}[t]
\begin{algorithmic}
\State $\mbox{\textbf{input:} example $x$, initial adversary $x'$,}$
\State $\quad \quad \quad \;\, \mbox{true class $c^{\dagger}$, classifier $f$}$
\State $\mbox{\textbf{output:} adversarial example $\tilde{x}$}$
\State $t = 0, x'_{t} = x'$
\While {TRUE} 
	\State $\tilde{x} = x'_{t}$
	\State $\mathcal{J} = \;$ MSE$(x, x'_{t})$
	\State $x'_{t} = \;$ ADAM$(x'_{t}, \mathcal{J})$
	\State $c'_{t} = \argmax[f(x'_{t})]$
	\State \textbf{if} $c'_{t} == c^{\dagger}$ \textbf{then} Break; \textbf{end}	
	\State $t = t+1$
\EndWhile \\
$\tilde{x} = x'_{t}$
\end{algorithmic}
\caption{DGM Adversarial Example Tuning}
\label{Alg:Ch2_DGM_adv_example_tune}
\end{algorithm}
\begin{figure*}[t]
\centering
\includegraphics[width=6.5in]{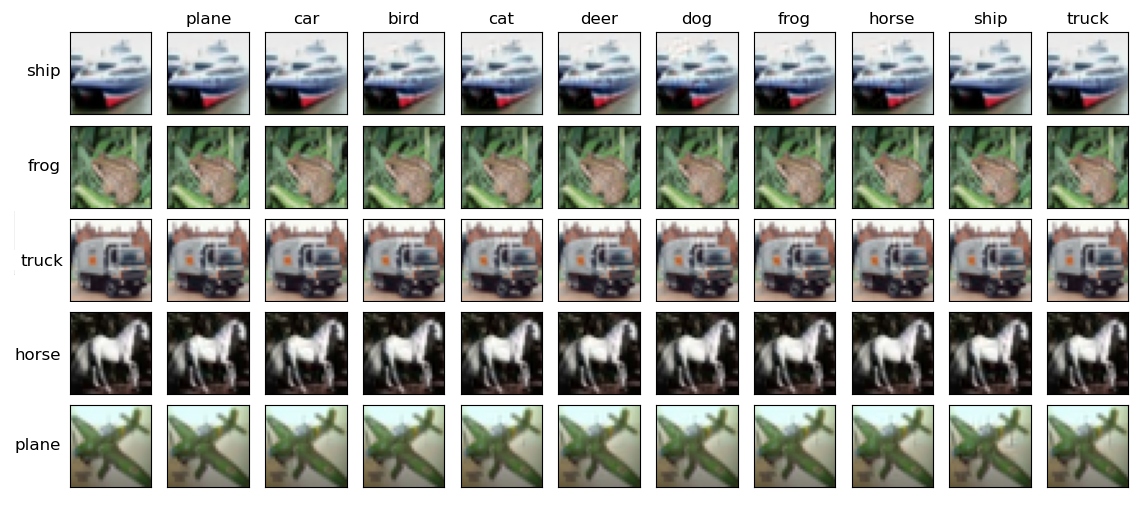}
\caption{Examples of targeted attack. DGM $l_{2}$ attack is applied to the CIFAR-10 dataset performing the targeted attack for each source/target pair. First column is the clean images} 
\label{Fig:Experiment_Targeted_CIFAR10}
\end{figure*}
\subsubsection{Adversary Tuning}
Attack method discussed in the previous section focuses on the effectiveness in finding an adversarial example $\tilde{x}$, which successfully leads to the misclassification with a high classification probability yet. One other objective of the attack is finding the minimum perturbation to make the adversarial example $\tilde{x}$ sufficiently close to the input example $x$. Input example can be seen as a fixed data point. Therefore, the goal of this optimisation is to find $\delta$ that minimises $l_{2}$ distance between $x$ and $\tilde{x}$. DGM solves such problem by formulating a simple objective of the iterative optimisation with Mean Square Error (MSE) as:
\begin{equation}
\text{min} \left( \text{MSE}(x, \tilde{x}) \right), \quad \text{such that} \; \tilde{x} \in [\text{min}(x), \text{max}(x)]^{m}
\label{Eq:DGM_tuning_objective}
\end{equation}
where $\tilde{x} = x + \delta$ is an adversarial example. To achieve the goal to find $\delta$ that minimises $\text{MSE}(x, \tilde{x})$, ADAM \cite{ADAM} optimiser is deployed in DGM method as detailed in Algorithm~\ref{Alg:Ch2_DGM_adv_example_tune}.

\subsubsection{Source Code}
The source code to reproduce experimental results can be found from 
https://github.com/jpark04-qub/SQBA